\documentclass[letterpaper, 10 pt, conference]{ieeeconf}  

\IEEEoverridecommandlockouts                              

\usepackage{graphicx} 
\usepackage{amsmath} 
\usepackage{amssymb}  
\usepackage{bm}
\usepackage{lipsum}
\usepackage{csquotes}
\usepackage{xcolor}
\usepackage{svg}
\usepackage{adjustbox}
\usepackage[noadjust]{cite}
\usepackage{url}
\usepackage{makecell} 
\usepackage[caption=false, font=footnotesize]{subfig}
\usepackage{etoolbox}
\usepackage{hyperref}
\usepackage{tikz}

\newtoggle{DETAILS}
\toggletrue{DETAILS}
\newcommand{\mytoggle}[2]{\iftoggle{DETAILS}{\color{blue}#1\color{black}}{#2}}
\newcommand{\detail}[1]{\mytoggle{#1}{}}

\newcommand{\vectau}{\bm{\tau}}
\newcommand{\vecx}{\bm{x}}
\newcommand{\vecu}{\bm{u}}
\newcommand{\vecp}{\bm{p}}
\newcommand{\veca}{\bm{a}}
\newcommand{\vecb}{\bm{b}}
\newcommand{\vecr}{\bm{r}}

\newcommand{\veczero}{\bm{0}}
\newcommand{\twist}{\mathcal{V}}

\newcommand{\norm}[1]{\left\lVert#1\right\rVert}
\newcommand{\ff}{{f\kern-0.15em f}} 

\DeclareMathOperator*{\argmin}{arg\,min}
\usepackage[short]{optidef}

\newcommand{\tikzcircle}[2][red,fill=red]{\tikz[baseline=-0.75ex]\draw[#1,radius=#2] (0,0) circle ;\hspace*{-0.1em}}%

\title{\LARGE \bf
  \detail{Extended Version of } GTGraffiti: Spray Painting Graffiti Art \\from Human Painting Motions with a Cable Driven Parallel Robot}

\author{
Gerry Chen, Sereym Baek, Juan-Diego Florez, Wanli Qian, \\Sang-won Leigh, Seth Hutchinson, and Frank Dellaert%
\thanks{This work was supported by the NSF under Grant No. 2008302}
\thanks{Institute for Robotics and Intelligent Machines, College of Computing,
Georgia Institute of Technology,
Atlanta, GA, \{gchen328, sereymbaek, jdfc3, wqian39, seth, fd27\}@gatech.edu}
}

\begin{document}

\maketitle
\thispagestyle{empty}
\pagestyle{empty}

\begin{abstract}

We present GTGraffiti, a graffiti painting system from Georgia Tech that tackles challenges in art, hardware, and human-robot collaboration.
The problem of painting graffiti in a human style is particularly challenging and requires a system-level approach because the robotics and art must be designed around each other.  
The robot must be highly dynamic over a large workspace while the artist must work within the robot's limitations.
Our approach consists of three stages: artwork capture, robot hardware, and planning \& control.
We use motion capture to capture collaborator painting motions which are then composed and processed into a time-varying linear feedback controller for a cable-driven parallel robot (CDPR) to execute.
In this work, we will describe the capturing process, the design and construction of a purpose-built CDPR, and the software for turning an artist's vision into control commands.
Our work represents an important step towards faithfully recreating human graffiti artwork by demonstrating that we can reproduce artist motions up to 2m/s and 20m/s$^2$ within 9.3mm RMSE to paint artworks.
\detail{Changes to the submitted manuscript are colored in blue.}

\end{abstract}

\section{Introduction and Related Work}\label{sec:intro}
Spray painting graffiti art in a human style is an important, open problem that requires a systems approach.
In this paper, we take the first step towards creating a system that can capture human graffiti artwork and collaborate with artists to create new and copied artworks in the public settings that define graffiti.
In addition to the well-established sociological motivations for reproducing graffiti art \cite{macdowall2006praise}, robot art is intrinsically motivating for its marriage of art and technology.
By possessing physical abilities beyond those of its collaborating artists, a graffiti robot could reveal new artistic avenues highlighting human-robot collaboration for disabled \cite{Liekens20youtube_disabledKennyCableGrafffiti} and able-bodied artists alike.
To act as the hand of an artist poses inherently interconnected problems in art, hardware, control, and human-robot interaction.
Furthermore, the technology required to produce the large-scale, dynamic motions required for graffiti has applications in  warehouse/industrial logistics \cite{Gouttefarde15TRO_CableControl}, agriculture \cite{Pagan18_cable_RTK}, construction \cite{Shahmiri16_cable_survey, bostelman94isrm_nist_robocrane}, and motion simulation \cite{Miermeister16iros_CableRobotSimulator}.
Creating graffiti art with a robot requires (1) capturing the motions of artists painting, (2) creating a robot that can achieve comparable motions to human artists, and (3) implementing algorithms that would allow the robot to execute on the artists' visions.  Despite considerable progress in each of these tasks, to our knowledge, no system has been demonstrated to achieve all three.

\begin{figure}
    \centering
    \includegraphics[width=\linewidth]{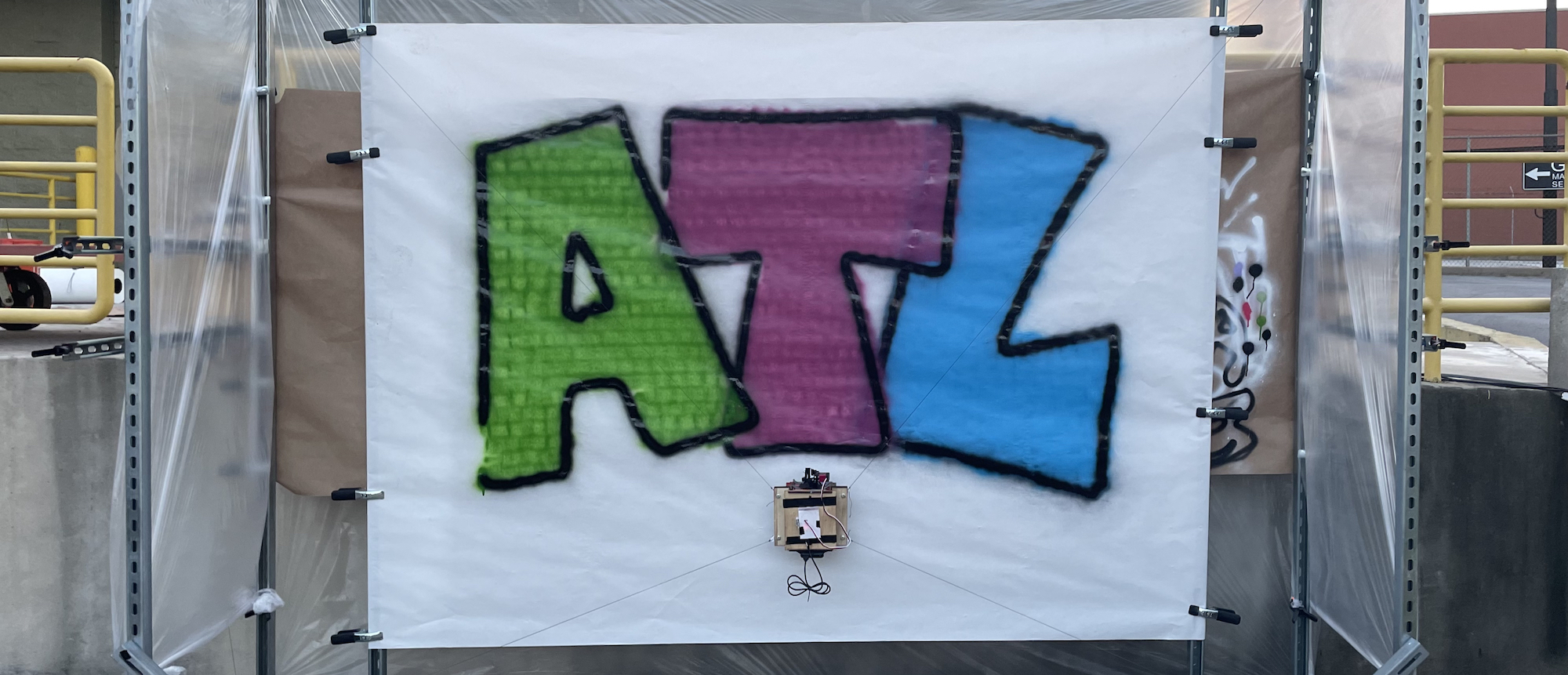}
    \caption{
    Our system captures artist painting motions of individual letter outlines which are composed and processed into controls for a cable robot to execute.  Our system produced this painting of the letters ``ATL'' (Atlanta).
    }
    \label{fig:GT_img}
    \vspace*{-1em}
\end{figure}

Prior work exists in capturing graffiti art, most notably the Graffiti Markup Language (GML) project \cite{Wilkinson21website_GML_000000book}.  Although the project has been successful in generating a large library of graffiti artwork, almost all the data was captured from digital interfaces (e.g. stylus) rather than full-body painting motions.
This is problematic because an artist's creative process may differ between virtual and physical mediums and because ignoring the physical painting motions neglects the challenge of generating robot trajectories.
An exception is the GML Recording Machine \cite{Yildirim11website_gml_recording_machine}, though it captures only 2 degree of freedom (DoF) planar motion.

Robots developed for spray painting have seen considerably more attention.
Serial arm manipulators and gantry-based systems are precise and mature, but arms do not scale well to large workspaces \cite{Scalera18sir_airbrushModel,Berio16iros_dynamic_strokes} and gantry-based systems exhibit a tradeoff between size and portability \cite{Scalera18jirs_watercolor,Roy21youtube_gantry}. 
Mobile manipulators address these issues, but are currently not as dynamic or precise as human artists \cite{Jun16iros_Humanoid_Graffiti}.  
Aerial robots are popular for their ability to paint otherwise inaccessible walls, but have been cited as being difficult to accurately control due to susceptibility to disturbances and comparatively limited acceleration capabilities \cite{Uryasheva19siggraph_multiDroneGraffiti,Vempati18ral_paintcopter,Galea16expr_stipplingDrone,Tsaru18web_droneMuralGraffiti}.
Cable-based systems appear to be promising,
but so far \cite{Albert21web_commertialproduct,Liekens20youtube_disabledKennyCableGrafffiti} have only demonstrated raster- or stippling- style painting while \cite{Lehni08perspecta_graffitibot_hektor} has not demonstrated the highly dynamic motions employed by human artists.


Finally, despite prior research in robot control and artistic composition, the software to enable graffiti painting does not currently exist.
CDPR control (further discussed in Section \ref{sssec:CDPR_lit_review}) is relatively well understood, but has not been demonstrated for dynamic graffiti trajectories.
Research on industrial painting robots has thoroughly studied paint dispersion and trajectory generation, but is primarily concerned with uniform coats on curved surfaces in contrast to graffiti art's non-uniform coats on flat surfaces \cite{Chen17ir_paintThicknessReview,Chen09ir_paintPlanningReview,Andulkar15jms_paintTrajGen,Zeng14ijca_sprayPaintTrajOpt_manyTimes}.
Berio is notable for his research in graffiti composition and stylization \cite{Berio19expr_graffitiLayering,Berio16iros_dynamic_strokes,Berio17mc_stylisation,Berio15wca_graffiti}, but focuses on digital rendering as opposed to producing trajectories. 

\begin{figure}
    \centering
    \includegraphics[width=\linewidth, trim=0 375 0 0, clip]{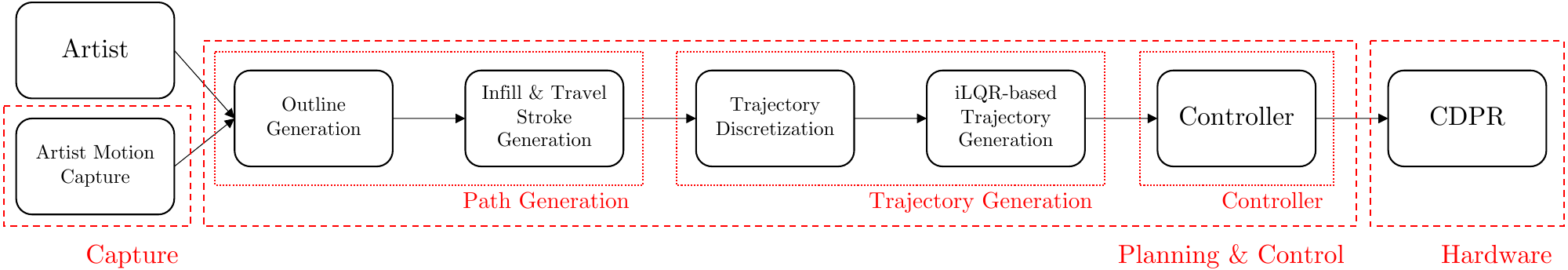}
    \caption{This system overview depicts the capture, hardware, and planning \& control components of our system.}
    \label{fig:system_overview}
\end{figure}

We argue that the problem of creating graffiti artwork is sufficiently expansive and its components codependent that it requires a system-level approach.
In this work, we propose a novel system towards creating graffiti artwork by improving and coordinating the capture, hardware, and software requirements.
Figure \ref{fig:GT_img} depicts an example result of the GTGraffiti system summarized in Figure \ref{fig:system_overview}.
Our contributions include:
\begin{itemize}
    \item \textbf{capturing} a library of 6 DoF trajectories for creating graffiti artwork using motion capture (mocap),
    \item designing, building, and testing the \textbf{hardware} for a purpose-made robot platform to paint graffiti,
    \item proposing a \textbf{planning and control} pipeline to realize high-level artistic descriptions into motor torque commands, and
    \item demonstrating a system that can paint human-style graffiti artwork.
\end{itemize}

\section{Capture}\label{sec:record}
The capture process is important for both learning the motions to produce graffiti art and establishing robot capability requirements.
As such, our capture process is focused on obtaining the most artistically meaningful data while omitting less relevant data.
In this work, we collect a library of simple, composable shape outlines.

\subsection{Design Considerations}

We first capture artwork using an OptiTrack™ mocap system for its simplicity and accuracy.
Mocap systems have the advantage of directly outputting positions and orientations of rigid bodies with sub-millimeter accuracy which trivializes the process of obtaining the 6D trajectories of a spray paint can during painting.  
As we will discuss in Section \ref{sec:software}, the processing and rendering components of our pipeline can optionally use other forms of captured artwork such as Scalable Vector Graphics (SVG) and GML files in addition to mocap data.

We opt to capture only the outlines of shapes and omit the infills because, according to an artist collaborator, the particular pattern used to fill-in a shape is largely arbitrary and algorithmically generating one does not significantly detract from artistic value.   Furthermore, the easiest infill path for a human may not be the easiest for a robot.

\subsection{Approach}
\begin{figure}
    \centering
    \includegraphics[height=.4\linewidth]{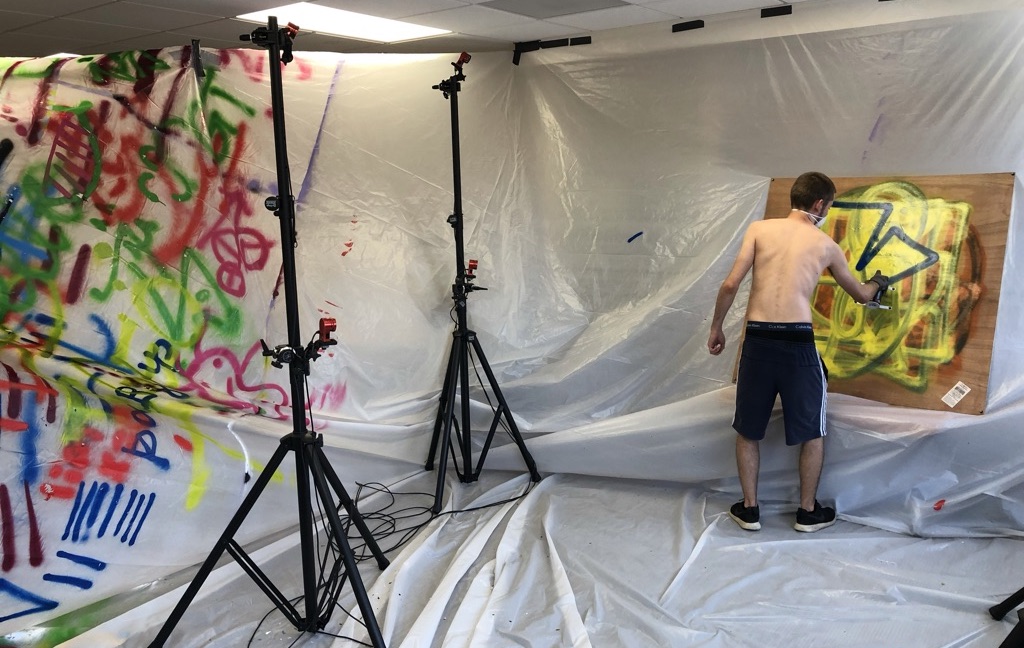}
    \vspace*{-.5em}
    \caption{Mocap setup for capturing painting motions of collaborating artists.}
    \label{fig:capture_spraycan}
\end{figure}

We collected the full 6D trajectories of the spray cans and painting surfaces (plywood sheets) as two graffiti artist collaborators painted.
    Four mocap position markers were affixed each to the can (as shown in Figure \ref{fig:capture_spraycan}) and painting surface to extract the 6DoF poses for each time step at 120Hz.  
For each art collaborator, the 26 letters of the English alphabet were captured along with special symbols such as punctuation marks and small doodles of the artists' choices (e.g. skull).  We convert the 6DoF pose of the spray can nozzle to the painting surface's reference frame using coordinate transformations.  Additional details can be found in \mytoggle{Appendix \ref{sec:app_cap}\vphantom{\cite{Chen21arxiv_gtgraffiti}}}{our accompanying arXiv paper \cite{Chen21arxiv_gtgraffiti}}.

To determine when the spray nozzle is being depressed (painting vs traveling motions), we applied a number of heuristics for each candidate painting motion segment including distance between start and end points (assuming outlines are closed curves), maximum speed, arc length, non-maximum suppression, and manual annotations.


\subsection{Results} \label{ssec:mocap_results}
\begin{figure} 
    \centering
    \vspace*{-.75em}
    \subfloat[]{%
        \label{fig:mocap_max}
        \includegraphics[height=0.4\linewidth,trim=260 85 260 180,clip]{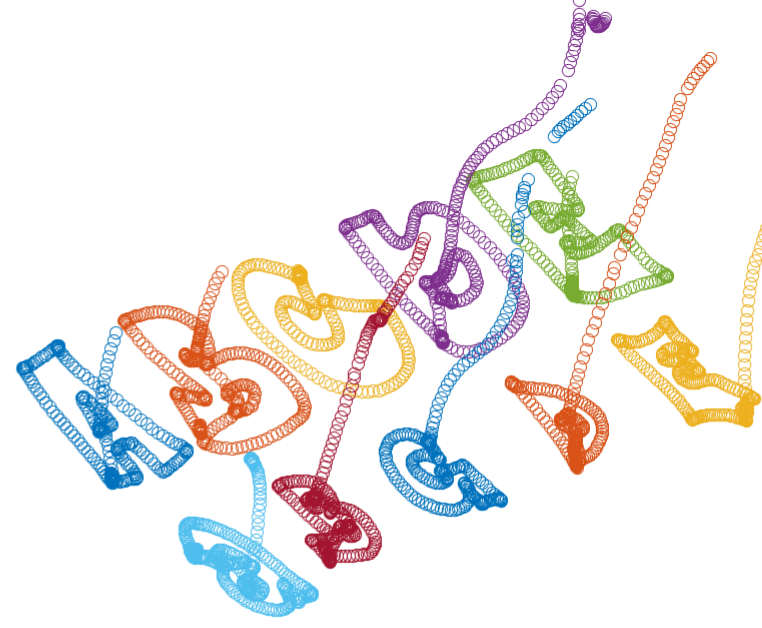}
    }
    \hfill
    \subfloat[]{%
        \label{fig:mocap_histograms}
        \includegraphics[height=0.4\linewidth]{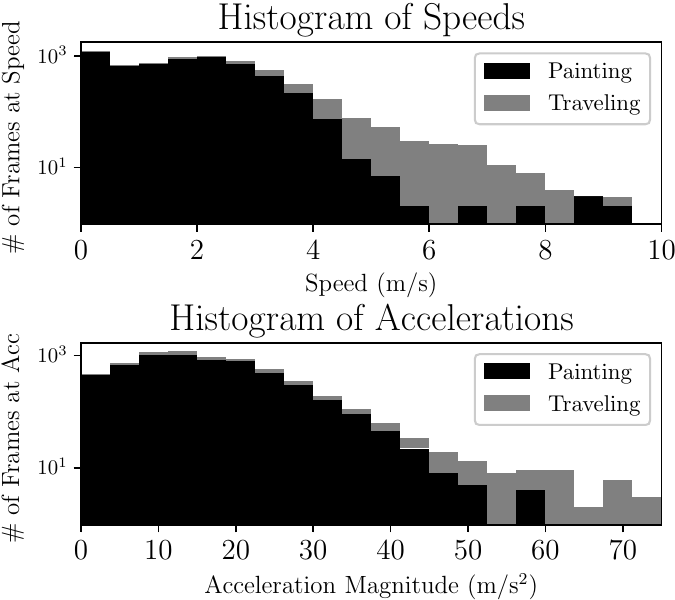}
    }
    \vspace*{-.25em}
    \caption{(a) Captured shape outlines for the letters ``ABCDE'' by our two collaborating artists.
    (b) Speeds and accelerations of a human graffiti artist during painting which help inform minimum requirements for the robot.}
\end{figure}

Figure \ref{fig:mocap_max} shows an example of our data by plotting the spray can nozzle translations. 
The fact that the data was collected while the artists were physically painting combined with the accuracy and 6DoF of motion capture gives our data the potential to better understand the nuanced motions of human graffiti painting e.g. biomechanically and with respect to can speeds, distances, and orientations.

We also study the speeds and accelerations reached by our collaborating artists during painting and traveling to inform the requirements of the cable robot.
From Figure \ref{fig:mocap_histograms}, we estimate a maximum speed of 6m/s and acceleration of 50m/s$^2$ to be sufficient to paint human-style graffiti artwork.

\subsection{Discussion \& Limitations}
Although motion capture's accuracy is unparalleled, there are several drawbacks.  Most notably, cost and setup hinder the accessibility and mobility of capture systems.  We were only able to capture motions in a controlled laboratory setting which limits the realism of the artwork. Additionally, human artists can move so fast that, even at 120Hz, our mocap system misses some detail.

The ability to capture nozzle actuation was also limited since we were unable to directly record actuation force which artists use to allow better paint control.  During data collection, an additional marker was actually placed on the tip of the artist's index finger to aid in identifying when the spray can nozzle was depressed.  Upon analyzing the data, however, this was not found to be a consistent method of annotating binary nozzle actuation let alone actuation force.
Even with various heuristics, manual annotation was needed to correct misclassifications in nozzle actuation.

Additional tags, characters, and full murals with photographs will be added to the library in the future.

\section{Robot Hardware}\label{sec:hardware}

Given the design requirements for painting graffiti based on human spray painting data, we believe that a CDPR is an ideal platform.  In this section we detail our robot hardware.

\subsection{Design Considerations} \label{ssec:hardware_design}
\begin{figure}
  \centering
  \includegraphics[height=0.30\linewidth, trim=0 110 0 110, clip]{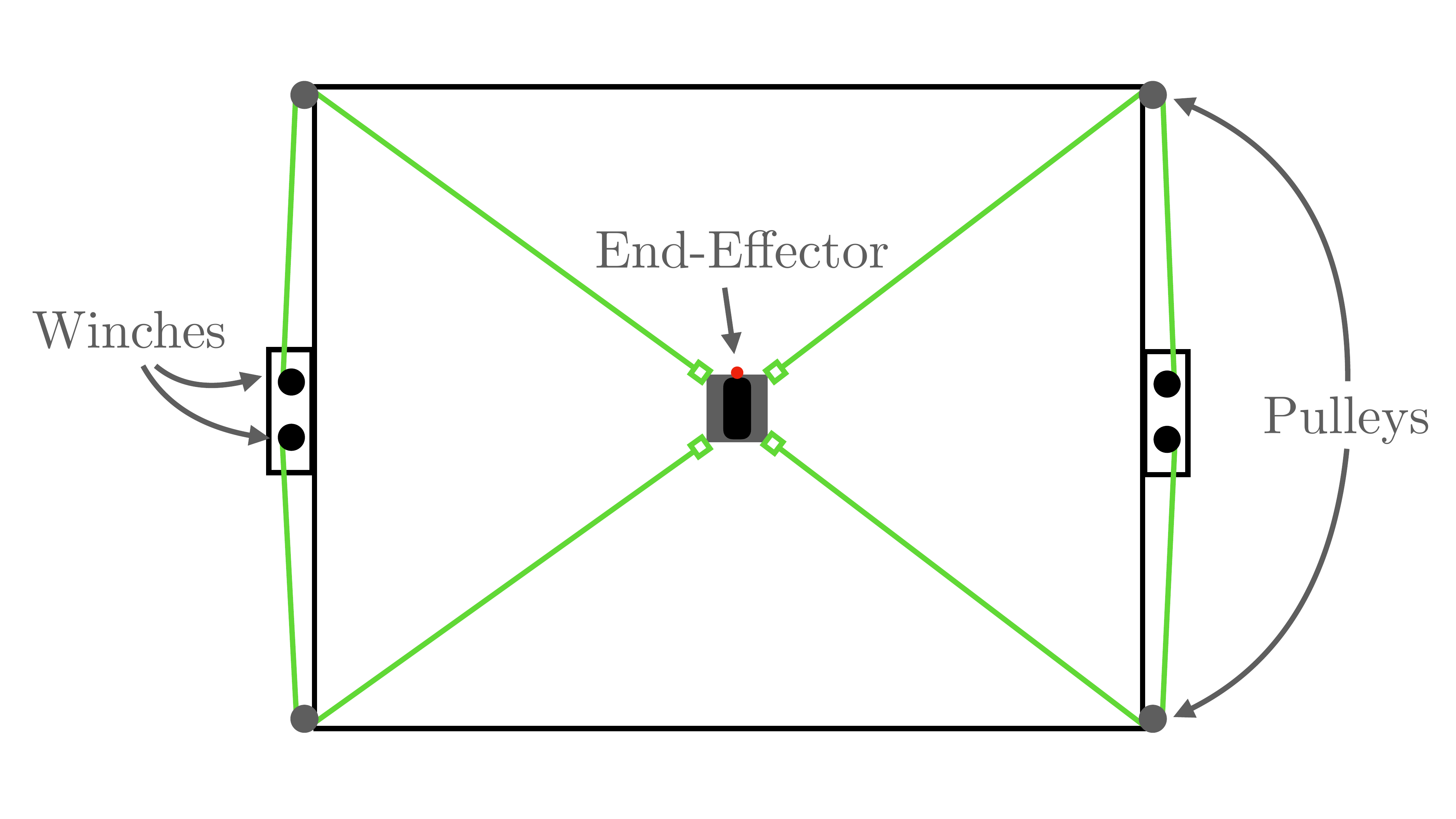}
  \caption{Our planar CDPR has a 4-cable, rectangular configuration with the end-effector in the center carrying the spray paint can.
  }
  \label{fig:cdpr_diagram}
\end{figure}
The primary design requirements for a graffiti painting system involve workspace size, maximum end-effector velocity, and maximum end-effector acceleration.  We seek a platform which can be scaled to a workspace 20mx20m or larger, though in this work we only seek a demonstration sized at a few meters.  Based on the analysis presented in Section \ref{ssec:mocap_results}, we determined that we require 6m/s and 50m/s$^2$ of speed and acceleration, respectively.  Assuming the mass of the spray can and actuating accessories do not exceed 2kg, including gravity the robot should be capable of exerting 120N upward and comparable forces in other directions.

Secondary design requirements include portability, accuracy, and stiffness.
It should be feasible to disassemble and reassemble the robot on-site at the wall of a building.
Accuracy and stiffness are considered secondary constraints because, compared to art forms such as brush painting or sculpture, graffiti is less sensitive to positional inaccuracies and experiences less reaction force.
Based on the thickness of a line painted with a ``needle'' nozzle 5cm from the painting surface, we estimate 2.5cm of repeatability to be sufficient.  We estimate an accuracy of 1\% the size of the painting to be sufficient, based loosely on \cite{Valle56psychol_perceptualThresholds}.
We estimate external disturbances to be negligible based on paint reaction forces and historical Atlanta wind speeds.

CDPRs present ideal platforms for graffiti painting given the aforementioned requirements.  A CDPR is a robot whose end-effector is pulled by a set of cables which are driven by winches on a fixed base.  Due to properties of cables, CDPRs can scale to extraordinary sizes and speeds \cite{Nan11ijmp_FAST_telescope, Saptarshi20nasa_lunartelescope}, albeit with reduced stiffness.  These qualities make them ideally suited to the large but relatively undisturbed setting and modest accuracy requirement of graffiti painting.

CDPRs also have an active research community which has solved many challenges in workspace analysis \cite{Bosscher06TRO_wrenchFeasible, Bouchard09_WFW, Gouttefarde11TRO_cableWFW}, control, and estimation (further discussed in Section \ref{sssec:CDPR_lit_review}).
Preliminarily, based on \cite{Gouttefarde15TRO_CableControl}, we estimate a 1kHz update frequency to be necessary for real-time control.

Finally, we define the requirements to actuate the spray can nozzle.
For a full can of Montana BLACK 400mL, the force required to depress the nozzle was measured to be 27N and the displacement was measured to be 2mm.  Other 400mL spray cans by the brands Montana, Hardcore, and Kobra were found to have similar actuation forces and displacements.

\subsection{Approach}
Our CDPR uses 4 cables in a planar configuration to exert pulling forces on the end effector via 4 motor-driven winches (see Figure \ref{fig:cdpr_diagram}).
The end effector was built to be lightweight and carry the spray can and actuating electronics.
It has 4 mounting points to connect to the 4 CDPR cables.
The spray can nozzle actuating mechanism is wireless, battery-powered, and implemented using a servo with the lever-arm mechanism from \cite{Liekens20thingiverse_3dprint_spraymount}.  Complete design details can be found in \mytoggle{Appendix \ref{sec:app_hardware}}{our accompanying arXiv paper \cite{Chen21arxiv_gtgraffiti}}.

\subsection{Results}
\begin{figure}
    \centering
    \includegraphics[height=.32\linewidth]{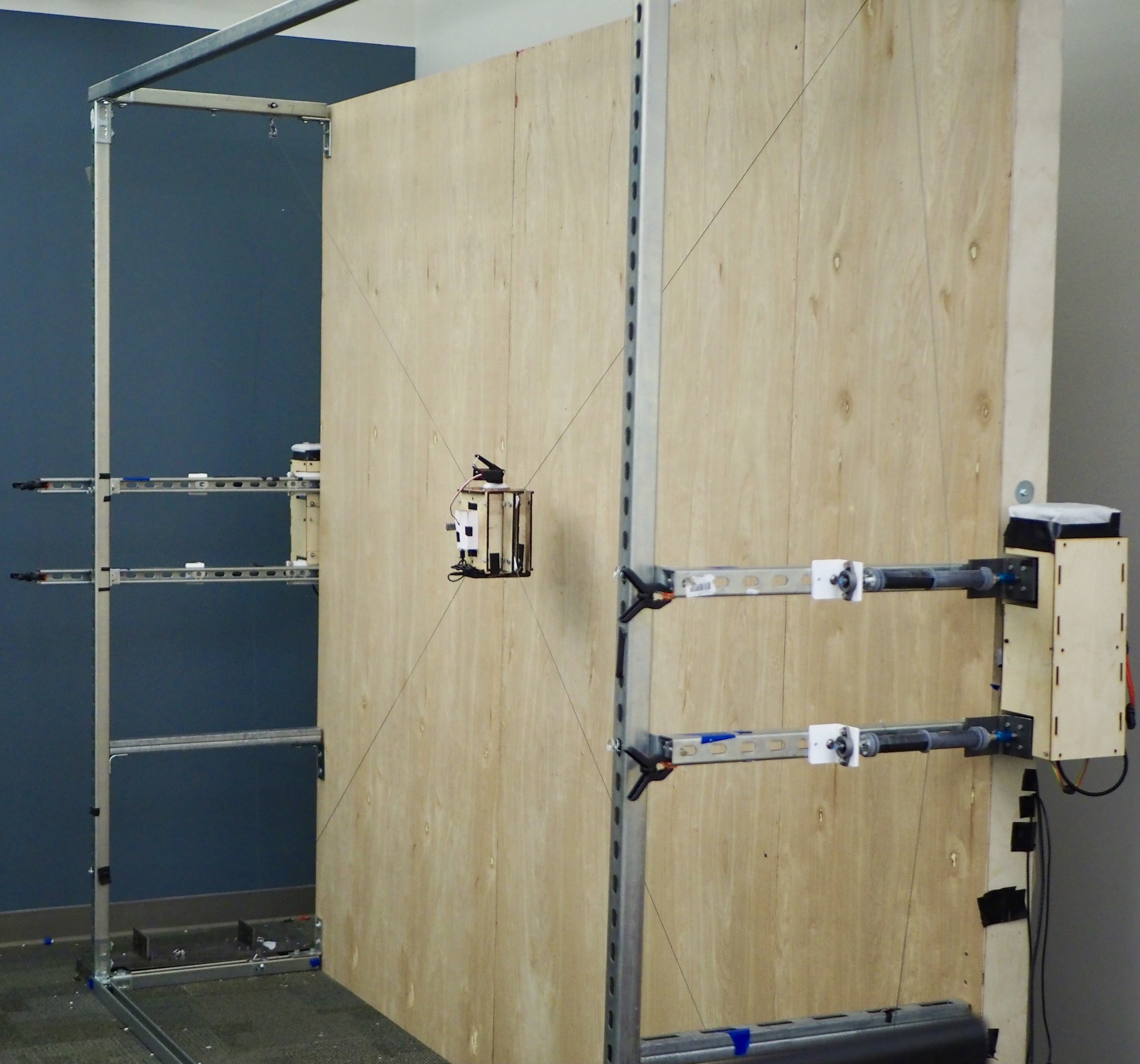}
    \includegraphics[height=.32\linewidth]{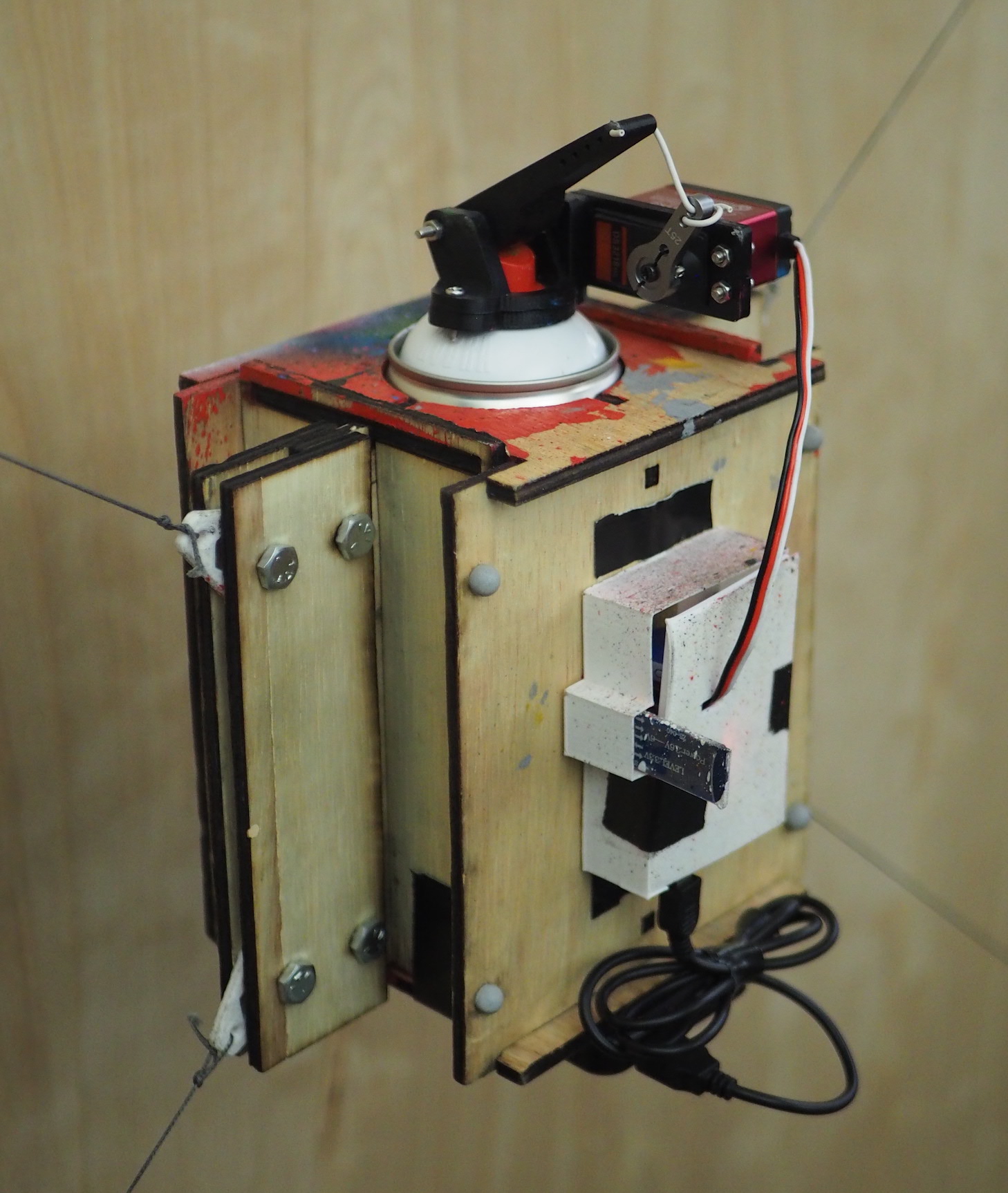}
    \includegraphics[height=.32\linewidth]{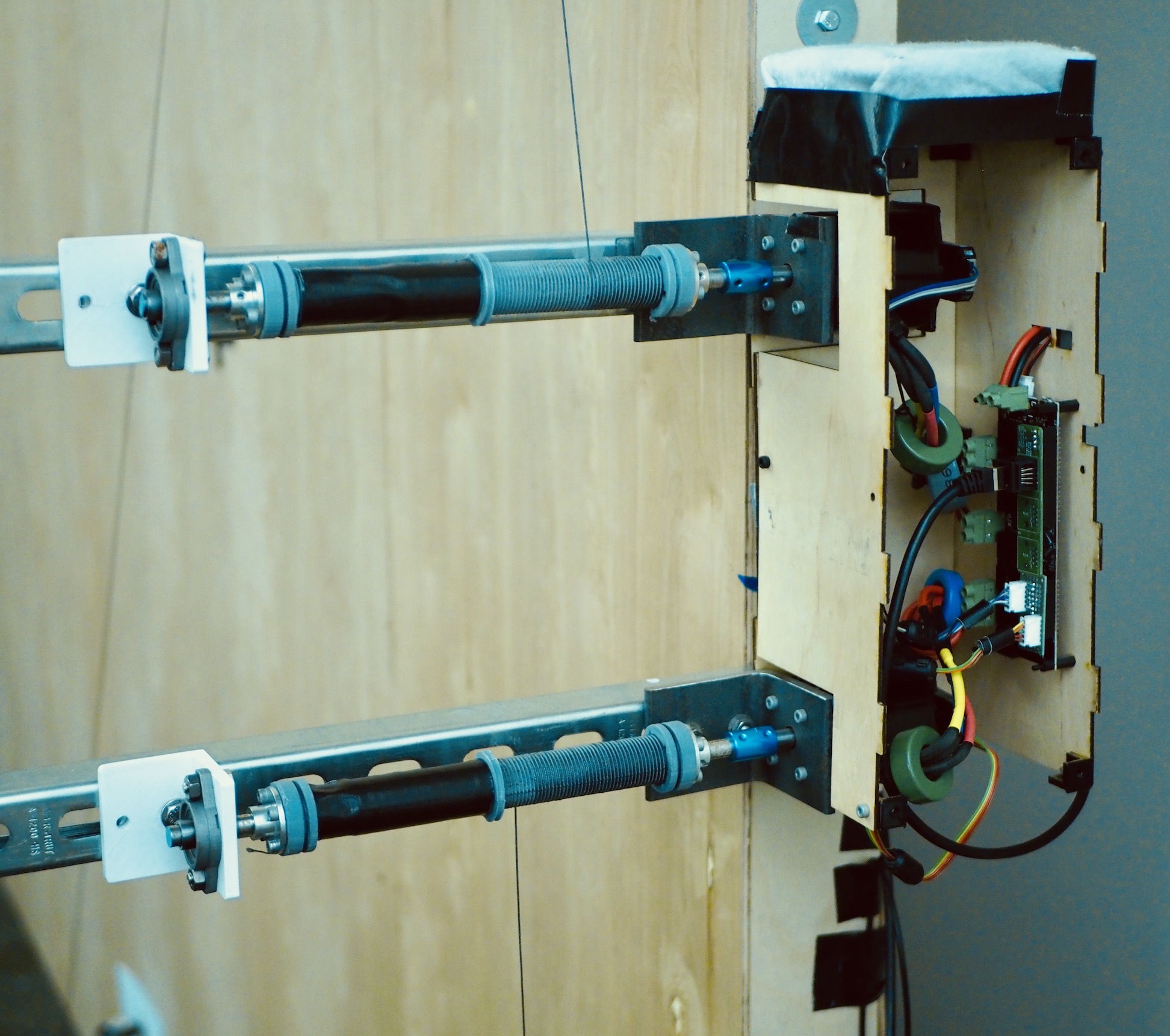}
    \caption{Our cable robot (left) includes an end effector that carries the spray paint and actuator electronics (center) and 4 winch assemblies, each consisting of a shared motor controller, motor, and helical winch (right, x2).}
    \label{fig:can_actuator}
\end{figure}

Our assembled robot is pictured in Figures \ref{fig:GT_img} and \ref{fig:can_actuator}.
The winches satisfy our design requirements with each being capable of pulling a 2kg mass on the cable up to 7.6m/s and 94m/s$^2$ and bidirectionally communicating at 1kHz.
The end-effector and spray can actuating mechanism are also pictured in Figure \ref{fig:can_actuator}.  The total mass varied between 1006g and 1317g depending on the spray can.
The spray can actuating mechanism was able to successfully depress the spray nozzle 100\% of the time in a trial of 100, 1 second long actuations.  The latency from commanding to dispensing paint was measured to be 400ms.

\subsection{Discussion \& Limitations} \label{ssec:mech_limitations}
We are able to paint well, as in Figure \ref{fig:GT_img}, despite not being able to use our 6DoF captured data to its full potential since we are limited to planar motion.  Simultaneously, we will discuss in Section \ref{sssec:software_results_traj} that the paint limits us to a maximum speed far below what the hardware is capable of. 
A combination of hardware upgrades and intelligent paint modeling and optimization are likely necessary to leverage our system's full potential, with actuation to move the nozzle closer to the canvas being paramount.


\section{Planning and Control}\label{sec:software}

From an artist's input, we must control the robot to paint.  We use a hierarchical approach with 3 levels:

\begin{enumerate}
    \item Path Generation: turn the artist's vision into a mathematical description
    \item Trajectory Generation: find a trajectory within the robot's capabilities while respecting the artist's vision to the maximum extent possible
    \item CDPR Control: execute the trajectory online
\end{enumerate}

The interplay between the trajectory generation and CDPR control merits a summarized precursor explanation for clarity.  During the trajectory generation phase, the optimal control problem of tracking a desired trajectory is solved \emph{offline}.  The iterative Linear Quadratic Regulator (iLQR) method \cite{Todorov05acc_iLGQ} -- iterating by applying the linearized system and control law forward in time to obtain a new linearized feedback law -- is used to solve the optimal control problem.  The feedback law from the final pass can then be used as the online controller.


\subsection{Design Considerations} \label{ssec:software_design}
\subsubsection{Path Generation}
In this work, the artist composes artworks using the shapes in the shape library as a first step towards more general artistic descriptions.
Given an artist's specification for 
the placements of shapes from the library of captured art, we seek to generate the \textit{paths}, in the form of B\'{e}zier curves, that the spray can must follow.  This is a system-level problem because it requires suitable captured data, well-planned and modeled robot capabilities, and clear artist desires.
We divide path generation into (1) outline, primarily a human-robot interaction problem, (2) infill, a coverage path planning problem, and (3) travel, a problem of filling in discontinuities.  The latter two are unique to our system approach because, recalling the reasoning from Section \ref{sec:record}, only outlines are captured for the shape library.

\subsubsection{Trajectory Generation}
To create a physically realizeable trajectory that is as similar as possible to the artist's vision, we first discretize the path at 100Hz to obtain a direct-from-artist trajectory, $\vecx_d$ (within speed and acceleration limits), 
then apply an offline iLQR-based optimization to obtain a smoothed reference trajectory, $\vecx_\ff$, control signal, $\vecu_\ff$, time-varying feedback gains, $K$, and paint timing.


\label{sssec:iLQR_design}
Loosely inspired by \cite{Berio17mc_stylisation}, the iLQR-based optimization is used to strike a balance between the artist's intent and the ease of controlling the robot. 
We express the iLQR problem in discretized form with time index $k$ as:\vspace*{-0.2em}
\begin{argmini!}|s|
  {\vecx, \vecu}{\hspace*{-.2em}\sum_{k=0}^T \tilde{\vecx}[k]^TQ\tilde{\vecx}[k] + \tilde{\vecu}[k]^TR\tilde{\vecu}[k]\label{eq:iLQR_objective}}{\label{eq:iLQR}}{\vecx_\ff, \vecu_\ff\hspace*{-.1em} =\hspace*{-.3em}}
  \addConstraint{\vecx[k+1] = f(\vecx[k], \vecu[k])\label{eq:iLQR_constraint}}
  \addConstraint{\vecx[0]=\vecx_0\label{eq:iLQR_initial}}
\end{argmini!}
where 
$\tilde{\vecx}[k]\coloneqq \vecx[k]-\vecx_{d}[k]$ is the deviation of the state $\vecx[k]$ from $\vecx_d[k]$: the artist's intended trajectory, 
$\tilde{\vecu}[k]=\vecu[k]-\vecu_m$ is the deviation of the control $\vecu[k]$ from $\vecu_m$: the average of the minimum and maximum allowable torques \cite{Pott09ckin_forcedist_closedform,Miermeister12gcr_cdpr_autocalib}, 
$\vecx_\ff$ and $\vecu_\ff$ are the smoothed reference (nominal/\underline{f}eed\underline{f}orward) state and control signals, 
$Q$ and $R$ are the state objective and control cost matrices, 
$f(\vecx[k], \vecu[k])$ defines the nonlinear system dynamics, 
and $\vecx_0$ is the initial state.
The state consists of the cartesian position and velocity: $\vecx = \begin{bmatrix}p_x & p_y & \dot{p}_x & \dot{p}_y \end{bmatrix}$, where $\vecp$ denotes the position of the spray can's nozzle.  The control $\vecu$ consists of the four motor torques.  Section \ref{sssec:software_results_control} experimentally justifies why orientation is omitted.

\begin{figure}
  \centering
  \includegraphics[width=\linewidth,trim=60 0 0 0,clip]{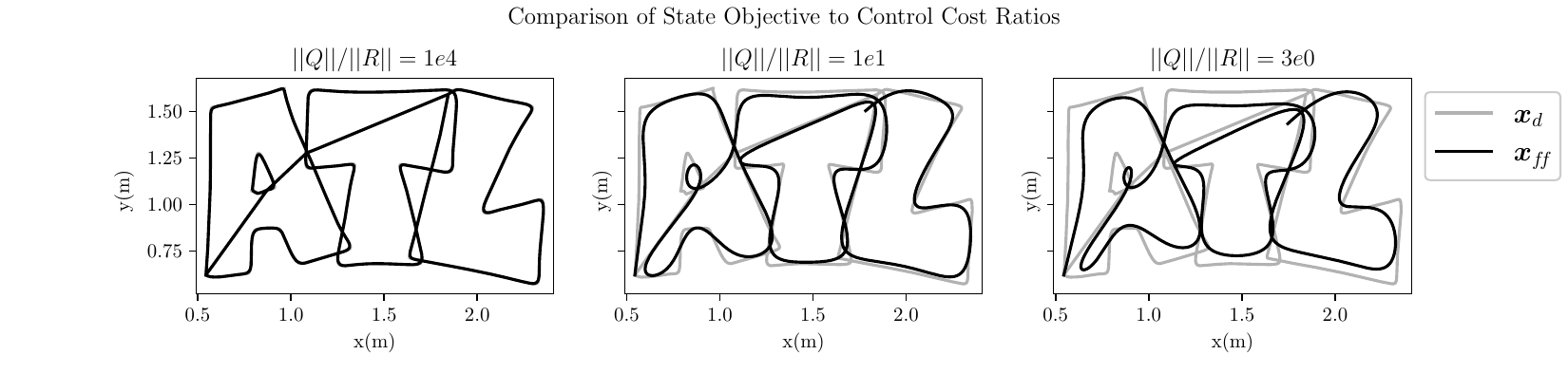}
  \caption{Stylization from iLQR Q vs R also observed by \cite{Berio17mc_stylisation}}
  \label{fig:iLQRstylization}
\end{figure}

Intuitively, the state objective matrix, $Q$, advocates for the artist and the control cost matrix, $R$, penalizes being near torque limits.
Interestingly, as discovered by \cite{Berio17mc_stylisation}, the relationship between $Q$ and $R$ can also be interpreted as an artistic parameter as visually depicted in Figure \ref{fig:iLQRstylization}.


\subsubsection{Control} \label{sssec:software_constraints_control}
We seek a controller that can control the cable robot to achieve motions comparable to a graffiti artist (requirements are the same as in Section \ref{ssec:hardware_design}). 

\label{sssec:CDPR_lit_review}
Our cable robot controller is inspired by prior works.
CDPR control places emphasis on ``tension distribution'' (TD) which is analogous to redundancy resolution in serial manipulators \cite{Miermeister16iros_CableRobotSimulator, Pott09ckin_forcedist_closedform, Agahi09tcasme_redundancyresolution_tension-velocity, Hassan11tro_tension_analysis, Taghirad11tro_redundancyres_3cases, Lamaury13ICRA_CableControlFF, Gouttefarde15TRO_CableControl,Shang20mmt_dualspace_CDPR}.
These approaches typically use or assume a
feedback controller whose control variable is a task space wrench.  The TD algorithm then computes the optimal motor torques (or cable tensions) required to achieve the desired task space wrench.  



\begin{figure}
  \centering
  \includegraphics[width=.55\linewidth, trim=0 175 0 0, clip]{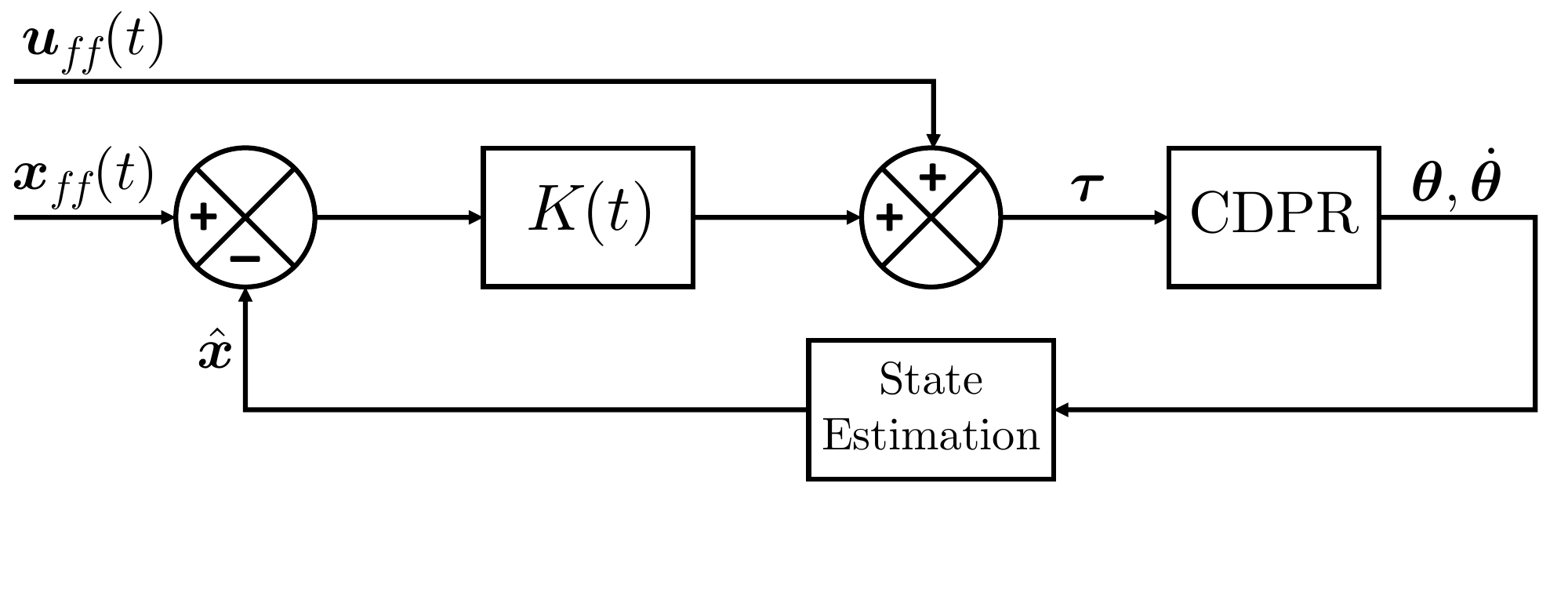}
  \caption{\hspace*{-.2em}CDPR controller block diagram, where $u_\ff$, $x_\ff$, and $K$ are precomputed offline using an iLQR-based optimizer implemented in GTSAM. 
  }
  \label{fig:blockdiagram}
\end{figure}

However, since we are using an iLQR-based optimizer, which produces locally optimal control laws as described in Section \ref{sssec:iLQR_design}, most aspects of control (including TD) have been shifted offline.  Our online controller is then a simple linear feedback controller.
Figure \ref{fig:blockdiagram} depicts a block diagram of our controller.  
Mathematically, our controller can be expressed as:
\begin{align}
  \vectau &= K(t) (\vecx_\ff(t) - \hat{\vecx}) + \vecu_\ff(t) \label{eq:online_controller}
\end{align}
where $\vectau$ is the 4-vector of motor torques, $\hat\vecx$ is the measured state, and $K(t)$ is the 4x4 time-varying gain matrix.

To compute $\hat{\vecx}$, we need to estimate the position and velocity of the spray can well enough to achieve our repeatability and accuracy requirements.  Nonlinear least squares solvers are commonly used for CDPR state estimation, but we will show that a simpler solution is sufficient for our application.

\subsection{Approach} \label{ssec:software_approach}
\subsubsection{Path Generation}
When generating the outline, the largest challenge involves communicating the artist's intent and robot's capabilities between the artist and computer.
In this work, we apply constraints to the artist when they are specifying their artistic vision.
To constrain the canvas size, we use a rectangular approximation of the wrench feasible workspace (WFW) \cite{Bouchard09_WFW,Gouttefarde11TRO_cableWFW} to define the space in which the artist may place library objects.
To constrain layering specifications, we impose a strict layering of shapes such that each shape is either entirely on top of or entirely beneath another.
Due to our robot's planar limitation, we project the nozzle position for each frame from the mocap data onto the painting surface to form an ordered set of 2D line segments (allowing us to retain velocity information to be used during trajectory generation).
Data sources other than our mocap library can be used but require velocity data.



To infill the shapes we apply an exact cellular decomposition and use a standard horizontal ``zigzag'' path within each cell \cite{Galceran13ras_coveragePathPlanning_survey}.
Similar to the way each artist chooses a strategy for infilling based on personal preference (according to our artist collaborators), we choose this pattern for its ease of implementation and actuation: in most configurations our robot has the best control authority horizontally.
Further details on the way we decompose the infill and compute the exhaustive walk to reduce nozzle actuation cycles are provided in \mytoggle{Appendix \ref{ssec:app_pathgen}}{our accompanying arXiv paper \cite{Chen21arxiv_gtgraffiti}}.
For each object, we paint the infill in the face color then the outline in the outline color.
Instead of applying a hidden line removal algorithm \cite{Devai86scg_hidden_line_elimination}, we simply finish painting each object before the next is started.

Finally, travel strokes must be added to make the path continuous in position.  Although making the paths continuous in direction is an option (as in \cite{Lehni08perspecta_graffitibot_hektor}), we opt to allow discontinuous directions but enforce continuous velocity in the trajectory generation stage.  For every pair of strokes, we add a straight line from the end of the previous stroke to the start of the next stroke if they are not already coincident.

\subsubsection{Trajectory Generation}
Unique to the system-level approach, we discretize outlines and infill/travel strokes differently due to the different ways the paths were generated.

For outlines, we have velocities from mocap data so we need to apply time-scaling.
We compute the original path's speed and acceleration using finite differences assuming each line segment takes 1s, then apply the linear transformation, $t' = ct$, to match a predefined maximum speed and/or maximum acceleration and sample $\vecx(t')$ from the path at 100Hz.  We choose limits of 1.2m/s and 20m/s$^2$ based on the spray paint dispersion described further in Section \ref{ssec:software_results}.

For the infills and travel strokes, we need to generate rest-to-rest trajectories with continuous velocities.  We choose trapezoidal velocity profiles for their popularity in industrial applications \cite{Siciliano10book_modellingPlanningControl}  
with limits of 0.5m/s and 20m/s$^2$ (based on spray paint dispersion).

\begin{figure}
  \centering
  \includegraphics[width=\linewidth,trim=0 320 170 0,clip]{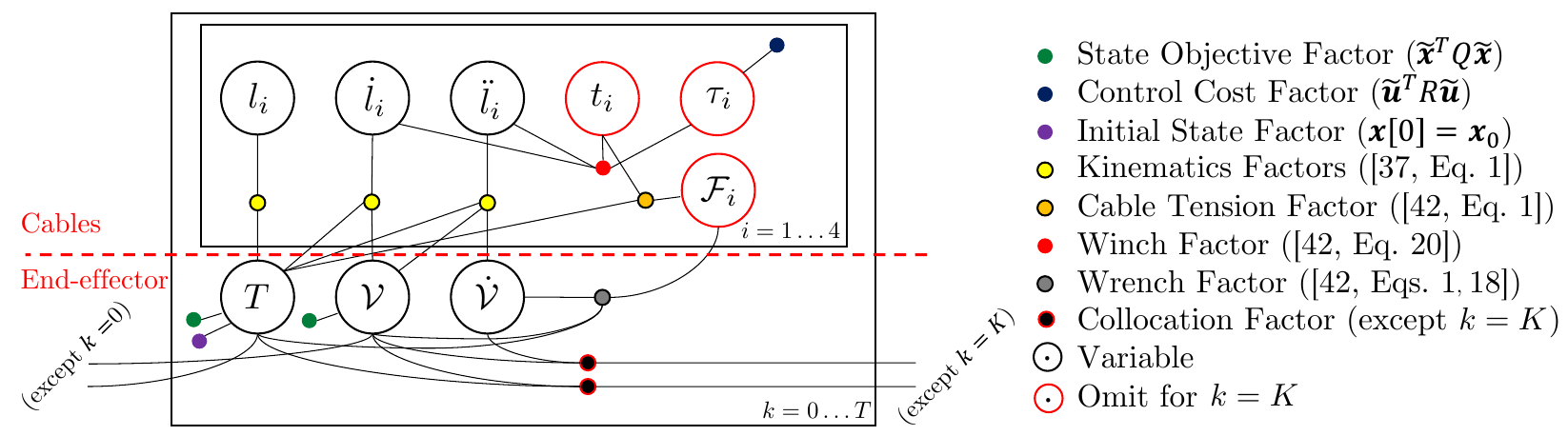}\vspace*{-.2em}
  \caption{Factor graph depicting the iLQR problem using plate notation.  Circles represent variables to be solved while dots represent objectives or equations.  $l_i$, $\dot{l}_i$, and $\ddot{l}_i$ represent cable length, speed, and acceleration respectively.  $t_i$, $\tau_i$, and $\mathcal{F}_i$ represent cable tension, motor torque, and the wrench on the end-effector caused by cable $i$, respectively.  $T$, $\mathcal{V}$, and $\dot{\mathcal{V}}$ represent end-effector pose, twist, and twist-acceleration, respectively. {\detail{See Appendix \ref{ssec:app_iLQR} for further details.}}}
  \label{fig:factorgraph}
\end{figure}

The iLQR-based optimization of \eqref{eq:iLQR} is performed offline using factor graphs and the GTSAM software library, but any iLQR implementation can be used. The system dynamics constraints \eqref{eq:iLQR_constraint} are drawn from prior works in CDPR control including the standard equations for kinematics and cable tension/wrench equilibrium \cite{Pott09ckin_forcedist_closedform}, winch model dynamics \cite{Lamaury13ICRA_CableControlFF,Gouttefarde15TRO_CableControl}, rigid body dynamics \cite{Lamaury13ICRA_CableControlFF}, and numerical integration \cite{Lau16IROS_CableSimulator,Butcher16chapter_eulerIntegration}. 
The iLQR problem is then expressed as the factor graph \cite{Dellaert17fnt_FactorGraphs,Yang21icra_eclqr,Chen19blog_lqr-blogpost} shown in Figure \ref{fig:factorgraph} and solved with the GTSAM software library using the Levenberg-Marquardt algorithm and variable elimination.  The solution gives $x_\ff$ and $u_\ff$, while the Bayes Net obtained by the final iteration's elimination step contains the locally optimal feedback gain matrix, $K$.  Full details on the equations and factor graph are available in \mytoggle{Appendix \ref{ssec:app_iLQR}}{our accompanying arXiv paper \cite{Chen21arxiv_gtgraffiti}}.

\subsubsection{Control} \label{sssec:software_approach_control}
First we interpolate $K$, $\vecx_\ff$, and $\vecu_\ff$ since the trajectory generation phase was discretized at 100Hz while the controller runs online at 1kHz.
$K$ and $\vecu_\ff$ are interpolated using a zero-order hold, while $\vecx_\ff$ is interpolated using a first-order extrapolation from the most recent discrete $\vecx_\ff$.


To estimate position we discard redundant information and to estimate velocity we use a linear least squares solution.
We discard the bottom two cable lengths then apply trigonometry using the top two cable lengths to estimate the spray can position assuming the spray can is always vertical.
To estimate velocity, we solve the linear least squares problem:
$
  \dot{\vecp} = \argmin_{\dot{\vecp}} \norm{\dot{l} - W^T \dot{\vecp}}_2^2 = W^{T+}\dot{l}
$
where $W$ is the \emph{wrench} matrix \cite{Lamaury13ICRA_CableControlFF} and $\cdot^+$ is the Moore-Penrose left inverse.

We also employ an \emph{offline} calibration whereby a rectangular trajectory is run while collecting mocap and robot log data.  A nonlinear least squares optimization is used to compute pulley locations and coefficients for cable length scaling which are hard-coded for subsequent runs.

\subsection{Results} \label{ssec:software_results}
\subsubsection{Path Generation}

\begin{figure}
  \centering
  \includegraphics[width=.3\linewidth]{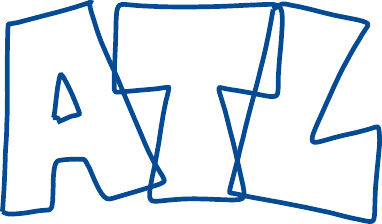}
  \includegraphics[width=.3\linewidth]{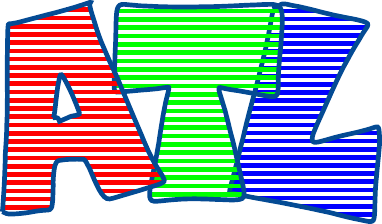}
  \includegraphics[width=.3\linewidth]{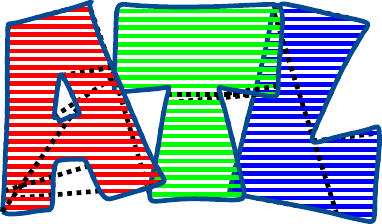}
  \caption{During path generation, we first produce the outline from artist inputs (left), then infill paths (center), and finally travel strokes (right).}
  \label{fig:path_gen}
\end{figure}
An example path generation result is shown in Figure \ref{fig:path_gen}.

\subsubsection{Trajectory Generation} \label{sssec:software_results_traj}
The speed and acceleration limits were tuned for our painting distance of around 12cm.
The outline limits of 1.2m/s and 20m/s$^2$ were tested using the Montana ``Skinny Cap Beige'' nozzle and the infill limits of 0.5m/s and 20m/s$^2$ were tested using the Montana ``flat jet cap wide'' nozzle.
Faster speeds resulted in incomplete coverage while slower speeds resulted in ``dripping''.

The offline iLQR-based optimization has $\mathcal{O}(n)$ complexity and runs at approximately half real-time (e.g. a 1-minute trajectory takes 2 minutes to optimize).  We chose $Q=\mathrm{diag}([1e4, 1e4, 0, 0])$ and $R=I_{4x4}$ as a balance between tracking accuracy and stability.

\subsubsection{Control} \label{sssec:software_results_control}

\begin{figure}
  \centering
  \includegraphics[width=\linewidth]{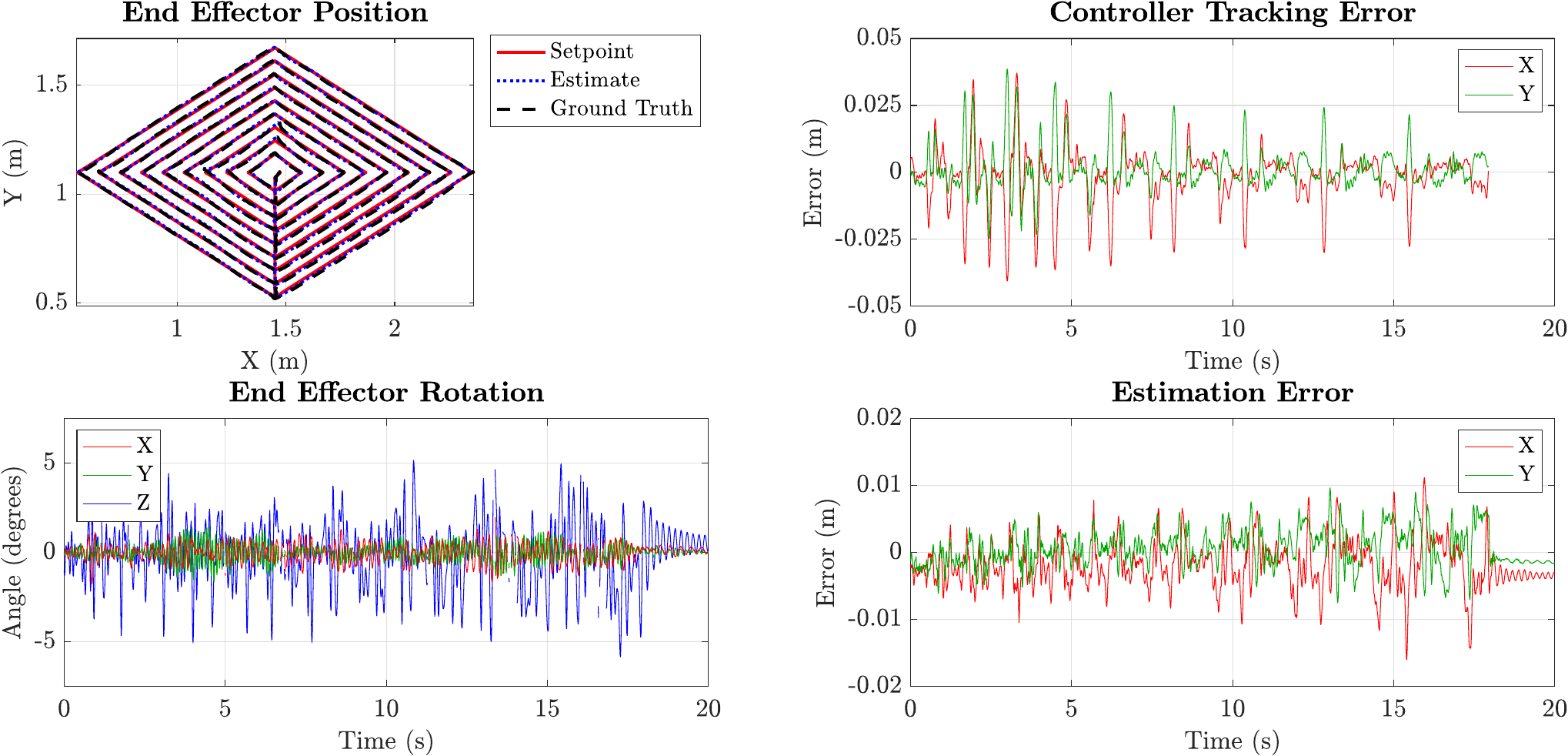}
  \caption{
  Setpoint $\vecx_\ff$, online estimate $\hat{\vecx}$, and ground-truth positions $\vecx$ of the spray can (top left), end effector rotation (bottom left), tracking error $\vecx_\ff-\hat{\vecx}$ (top right), and estimation error $\hat{\vecx}-\vecx$ (bottom right) for a challenging 2m/s, 20m/s$^2$ trajectory.
  }
  \label{fig:mocap_experiment}
\end{figure}

To evaluate our control stage, we use mocap for ground truth data and log $\vecx_\ff$ and $\hat{\vecx}$ from the robot at 100Hz for a challenging 2m/s and 20m/s$^2$ trajectory with sharp corners (Figure \ref{fig:mocap_experiment}).  The mocap and robot coordinate frames were aligned using the 4 pulley locations and the mocap data was piecewise cubic interpolated to match the 100Hz robot log frequency.  The control tracking's root mean square (RMS) error is 9.3mm and the position estimation's is 3.4mm.
We also validate our assumption that the end effector is always close to vertical, which is used both for online control and estimation, by measuring the RMS rotation deviation to be 0.45, 0.42, and 1.57 degrees in the horizontal, vertical, and normal directions respectively.
We believe that our proposed controller, which precomputes linear feedback gains offline, is accurate and easier to implement and useful for CDPR applications other than graffiti as well.







In addition to the painting in Figure \ref{fig:GT_img}, please refer to our supplemental video for additional painting results which qualitatively demonstrate our system's capabilities.


\subsection{Discussion \& Limitations} \label{ssec:software_limitations}
When specifying an artist's vision, the space of creative possibilities is immense.
Non-flat overlap topologies \cite{Berio19expr_graffitiLayering}, homographies, non-linear, and other 3D perspective transforms have artistic interest but are beyond the scope of this work.  Graffiti stylization \cite{Berio17mc_stylisation} and free-form inputs \cite{Berio18_expressive} are also beyond the scope of this work.  In return, the artists get to explore the maxim ``Creativity is born from limitations''.

Understanding the nuances of paint dispersion is another large area of study that is beyond the scope of this paper.  For example, the fact that human artists consistently paint solid lines at 6m/s yet our robot's lines begin losing complete coverage above 1.2m/s suggests a gap in our understanding.  We believe moving closer to the canvas (artists were on average 3.0cm$\pm$0.1cm from the painting surface vs 12cm for our robot) and modeling special effects such as flares, blending, and intentional dripping are logical next steps.


Although our controller is generally reliable and robust to modeling inaccuracies, we do find some limitations on the $Q$ matrix.  For large $Q$ matrices ($\norm{Q}_2/\norm{R}_2 \ge 1$e6N/m), the controller resonates with the natural vibrations 
of the cable robot 
causing instability, while for small $Q$ matrices ($\norm{Q}_2/\norm{R}_2 \le 1$e2N/m), the robot gets stuck for a few cm before overcoming the friction and returning to the setpoint trajectory.  Still, compared to other methods such as \cite{Pott09ckin_forcedist_closedform,Miermeister12gcr_cdpr_autocalib}, we believe the iLQR method to be easier because it requires less tuning and modeling effort.

\section{Conclusions \& Future Work}\label{sec:conclusions}
In this paper, we presented a system for painting human-style graffiti art.  Our work contributes to existing research by bridging three components in a systems approach: capturing artwork, building a graffiti robot, and planning and controlling the robot for painting graffiti.
We illustrated the co-dependencies of various system design choices which suggest that future research in robot art should consider a more holistic approach.  We also demonstrated that our system can successfully produce physical artworks from captured art.
Our work can be applied to graffiti preservation by recreating captured artwork, to human-robot collaboration in art by enhancing the physical capabilities of artists, and to other fields through technology transfer for large-scale dynamic motion.
Avenues for future work include a more portable graffiti capture device, better communication of robot limitations to the artist, style analysis and improvisation, paint dispersion analysis, real-time human-robot interaction, a larger workspace, and 6 DoF robot motion.







\section*{Acknowledgements}
We thank Max Ongena and Jules Dellaert for collaborating as graffiti artists, Prajwal Vedula and Zhangqi Luo for contributing to code, and Russel Gentry, Jake Thompkins, and Tristan Al-Haddad at Georgia Tech's Digital Fabrication Lab for their hardware assistance and allowing us to use their space for painting.

\detail{
\pagebreak
\appendices
\section{Capture}\label{sec:app_cap}
\subsection{Coordinate Frame Transformations}

The motion capture data is given in arbitrary ``world'' coordinates so we must convert the data into the painting surface's reference frame.
    The top left (y-axis), bottom left (origin), and bottom right (x-axis) markers of the painting surface are used to obtain the coordinate frame, ${}_wT_s[k]$, of the \underline{s}urface in the \underline{w}orld frame for each timestep, $k$, using Gram-Schmidt Orthogonalization for x then y, then using the cross product to obtain the z-axis.  Anecdotally, we found that the fourth mocap marker was never needed, though it would be useful in the event of one of the other three markers missing data.  For the spray can, a similar process is performed to obtain the \underline{c}an's frame, ${}_wT_c[k]$.  The pose of the spray can's \underline{n}ozzle in the spray can's frame, ${}_cT_n$, is obtained by manually measuring the position in the spray can's coordinate system and assigning the identity rotation.  
    For each timestep, $k$, the markers are used to obtain the coordinate frames of the \underline{s}urface, ${}_wT_s[k]$, and \underline{c}an, ${}_wT_c[k]$, both in the \underline{w}orld frame.  The pose of the spray can's \underline{n}ozzle in the spray can's frame, ${}_cT_n$, is obtained by manually measuring the position in the spray can's frame and assigning the identity rotation.  
Finally, the nozzle's pose in the painting surface's frame at timestep $k$, ${}_sT_n[k]$, can be expressed as
\begin{align*}
    {}_sT_n[k] &= ({}_wT_s^{-1}[k]) ~({}_wT_c[k]) ~ (_cT_n).
\end{align*}


\section{Robot Hardware}\label{sec:app_hardware}


\subsection{CDPR}

The cable mounting positions on the carriage and routing pulley locations are given in Table \ref{tab:cdpr_mountconfig}.

The frame is constructed from standard 12 gauge steel strut channel to dimensions 3.05m x 2.44m x 0.61m.  The four winches are 2.54cm in diameter with 1.5mm pitch helical grooves to drive 1mm Dyneema® (ultra-high molecular weight polyethylene) rope.  They are driven by 150kV D6374 motors from ODrive Robotics and controlled by two ODrive v3.6 56V motor drives.  The motor drives are connected via separate, isolated 1Mbaud CANbuses to a Teensy 4.0 microcontroller (MCU) running at 600MHz which runs the primary CDPR control (Section \ref{sssec:software_constraints_control}), sending torque commands and receiving angular position and velocity feedback to/from the motor drives.  The MCU also sends binary spray commands to the spray can actuator wirelessly with an HC-05 bluetooth module.

The motors are capable of 600 rad/s and 3.86Nm, which corresponds to 7.62m/s and 94.5m/s$^2$ for the 2.54cm diameter winch, the 1.96e-4 kgm$^2$ motor inertia, and a 2kg end-effector mass.  The communication via CANbus between the MCU and motor drives was measured to have a feedback-command round-trip latency of 626$\mu$s and easily achieves the required 1kHz update frequency.

\subsection{End Effector}

The end effector frame is composed of the six faces of a box plus two perpendicular midplanes parallel to the gravity vector to center the spray can. The 4 cables are mounted onto the midplane parallel to the painting surface via 1/4''-20 bolts and 3D printed hooks designed to fail before any other component.  The end effector was made from laser-cut 5mm hardwood.

The spray can nozzle actuating mechanism uses a hobby-grade ``20kg'' servo, 10,000mAh USB battery pack, Arduino Nano microcontroller, and HC-05 bluetooth module.  Note that large ($\ge$1000$\mu$F) bypass capacitors are required to prevent brownouts.

The mass breakdown of the 1006g--1317g end effector is given in Table \ref{tab:ee_mass}.  Note that the spray can varied from 113g to 424g depending on the fullness, brand, and part-to-part variability.


\begin{table}
  \centering
  \color{blue}
  \caption{CDPR cable configuration}
  \begin{tabular}{c|cc}
      \thead{Cable\\ Index} & \thead{End-effector Mounting\\ Location (m)} & \thead{Routing Pulley \\Location (m)}\\
      \hline
      1 & [0.094, -0.061, 0] & [1.52, -1.22, 0] \\
      2 & [0.094, 0.061, 0] & [1.52, 1.22, 0] \\
      3 & [-0.094, 0.061, 0] & [-1.52, 1.22, 0] \\
      4 & [-0.094, -0.061, 0] & [-1.52, -1.22, 0]
  \end{tabular}
  \label{tab:cdpr_mountconfig}
\end{table}

\begin{table}
  \centering
  \color{blue}
  \caption{End effector mass distribution}\label{tab:ee_mass}
  \begin{tabular}[]{c|r@{}l}
    Component & \multicolumn{2}{c}{Mass (g)} \\\hline
    Empty frame & 496& \\
    Battery & 231 \\
    Servo + lever-arm & 96&.0 \\
    Electronics & 70&.0 \\
    Spray can (max) & 424& \\\hline
    Total (max) & 1317&
  \end{tabular}
\end{table}

\subsection{Additional Discussions on Non-idealities}
\subsubsection{Parasitic Forces}
Friction (especially static) and motor cogging are hardware issues that plague CDPR control.  To reduce control difficulties, parasitic forces should be minimized where possible by ensuring smooth bearings and avoiding overpowered motors.  Perhaps counterintuitively, oversized motors can make CDPR control more difficult.  This is because motor cogging tends to scale with motor torque so not utilizing the full torque range of the motor causes the relative effect of cogging to be amplified.  For a similar reason, in optimal control formulations we do \emph{not} recommend using minimum energy objectives.  Instead, we recommend maximizing margins to torque limits (as we do in this work with \eqref{eq:iLQR_objective} and by defining $\hat\vecu[k]=\vecu[k]-\vecu_m$), using the ``centroid'' of the feasible polytope \cite{Gouttefarde15TRO_CableControl}, or a comparable method that encourages torques near the center of torque limits.

\subsubsection{Oscillations}
Out-of-plane oscillations, both translational and rotational, were rarely problematic in practice, especially when median torque was increased.  Interestingly, the end effector rotation data from Figure \ref{fig:mocap_experiment} indicates that in-plane rotation exhibited more error than out-of-plane rotation.  We believe this to be because out-of-plane error is passively stabilized to 0 by (approximately) spring-mass-damper dynamics.  In contrast, due to omitting the orientation of the end effector in our closed loop controller \eqref{eq:online_controller}, the center of oscillation for in-plane rotation is non-zero (dependent upon the $\vecu_\ff$).  For the application of painting graffiti, however, we found qualitatively that these errors were not problematic and often trumped by mis-alignment in the nozzle orientation (due to human error).

\subsubsection{Known Issue}
Finally, we would like to note that a hardware-related issue with the bottom-left winch's encoder (which could not be replaced prior to submission) resulted in visible oscillation artifacts in some paintings that increase from right-to-left.  We believe that the smooth behavior on the right side of the workspace is more typical and will be observed throughout the workspace after replacing the encoder.

\section{Planning and Control}\label{sec:app_control}
\subsection{Infill Path Generation}\label{ssec:app_pathgen}
\subsubsection{Cellular Decomposition}  We use horizontal sweep lines and form new cell boundaries anytime there is a change in the number of times the sweep line intersects the infill boundary.  Then, we traverse the cells from top to bottom, painting each one in a horizontal zig-zag pattern.  When multiple cells are at the same height, we choose the order according to the direction of the last painted line: if the last painted line was left-to-right then we traverse the next cells right-to-left (and begin the zig-zag pattern for each cell right-to-left) and vice-versa.  
Although more sophisticated coverage algorithms exist, we found that robot travel motions are orders of magnitude faster than paint color changes so inefficiencies in the infill paths are relatively insignificant from a speed perspective.  Improvements to reduce nozzle actuations would be desirable since nozzle actuation has greater latency and variability than CDPR motions.

\subsubsection{Composition}The order that we compose paths -- infill in the face color then outline in the outline color -- differs from the way human artists typically paint.
Human artists typically start with the outline in the face color, proceed to infill in the face color, then re-assert the outline in the outline color. The initial face color outline is usually for visual reference, but our system does not have such visualization constraints so we opt to omit the initial face color outline.

\subsection{Background: Factor Graphs for Optimal Control}
A factor graph is a bipartite graph consisting of \emph{variables} and \emph{factors} connected by edges \cite{Koller09book_pgm}.  Variables are unknowns to solve for while factors express joint probability distributions.  For optimal control problems, we can instead interpret factors as optimization objectives or constraint equations by taking the negative log likelihood of a probability distribution (with ``zero-covariance'' in the case of a constraint).  To optimize a factor graph, we can apply the \emph{variable elimination} algorithm which is roughly analogous to algebraically optimizing for a single variable then substituting it back into the remaining equations and repeating for each variable.  It can be shown that variable elimination on a factor graph with only Gaussian (``linear'') factors is exactly equivalent to the Dynamic Algebraic Ricatti Equation for solving discrete, finite-horizon LQR problems \cite{Yang21icra_eclqr,Chen19blog_lqr-blogpost}, and that GTSAM's Levenberg-Marquardt implementation for optimizing a factor graph with nonlinear factors is exactly equivalent to the implementation of iLQR described by \cite{Todorov05acc_iLGQ}.
For additional background, we refer the reader to \cite{Dellaert17fnt_FactorGraphs} for an introduction to factor graphs and \cite{Yang21icra_eclqr,Chen19blog_lqr-blogpost} for details on how they can be applied to optimal control with and without additional constraints, respectively.

\subsection{iLQR using Factor Graphs}\label{ssec:app_iLQR}
The iLQR problem is expressed as the factor graph in Figure \ref{fig:factorgraph}, where each equation in the iLQR constrained minimization problem \eqref{eq:iLQR} corresonds to one or more factors in Figure \ref{fig:factorgraph}:
\begin{itemize}
  \item \eqref{eq:iLQR_objective} is represented by the
  \tikzcircle[fill={rgb,255:red,0; green,127; blue,59}]{0.5ex} State Objective and 
  \tikzcircle[fill={rgb,255:red,0; green,32; blue,96}]{0.5ex} Control Cost factors; 
  \item \eqref{eq:iLQR_initial} is represented by the 
  \tikzcircle[fill={rgb,255:red,112; green,48; blue,160}]{0.5ex} Initial State factor; and
  \item \eqref{eq:iLQR_constraint} is represented by the
  \tikzcircle[black, fill={rgb,255:red,255; green,255; blue,0}]{0.5ex} Kinematics, 
  \tikzcircle[black, fill={rgb,255:red,255; green,192; blue,0}]{0.5ex} Cable Tension, 
  \tikzcircle[fill={rgb,255:red,255; green,0; blue,0}]{0.5ex} Winch, 
  \tikzcircle[black, fill={rgb,255:red,127; green,127; blue,127}]{0.5ex} Wrench, and 
  \tikzcircle[red, fill={rgb,255:red,0; green,0; blue,0}]{0.5ex} Collocation factors.
\end{itemize}

To solve using the iLQR-based solver, we consider SE(2) planar motion for the kinematics and dynamics (in contrast to the online feedback controller \eqref{eq:online_controller} which considers only planar translational motion\footnote{Errata: \eqref{eq:iLQR} should be corrected to reflect that the state includes rotation.  Rotation is only removed for the online controller}).  When representing $T$, $\mathcal{V}$, or $\dot{\mathcal{V}}$ as vectors, we use the convention that the orientation is the first element and the translation the latter two.
The equations for each of the factors is given in Table \ref{tab:factorequations} and come from standard equations in CDPR literature, with example references given in the legend of Figure \ref{fig:factorgraph}.

\begin{table}
  \color{blue}
  \begin{center}
  \caption{Equations for the Factors in Figure \ref{fig:factorgraph}}
  \begin{tabular}{ccc}
    Factor Name & Factor Equation/Expression \\ \hline
    \tikzcircle[fill={rgb,255:red,0; green,127; blue,59}]{0.5ex}
    State Objective & $\norm{ T - T_d }_{Q^{-1/2}}^2$ \\
    \tikzcircle[fill={rgb,255:red,0; green,32; blue,96}]{0.5ex}
    Control Cost & $\norm{ \tau_i - (\tau_{min} + \tau_{max}) /2 }_{R^{-1/2}}^2$ \\[0.3em]
    \tikzcircle[fill={rgb,255:red,112; green,48; blue,160}]{0.5ex}
    Initial State & $T[1] = T_0$ \\
      & $\mathcal{V}[1] = \mathcal{V}_0 $ \\[0.5em]
    \tikzcircle[black, fill={rgb,255:red,255; green,255; blue,0}]{0.5ex}
    Kinematics & $ l_i = \norm{\vecr_i}_2 $ \\
      & $\dot{l}_i = \left([\mathrm{Ad}_{[\veczero;-\vecb_i]}]\mathcal{V}\right)\cdot [\veczero;\hat{\vecr}_i] $ \\
      & $\ddot{l}_i = \left([\mathrm{Ad}_{[\veczero;-\vecb_i]}]\dot{\mathcal{V}} \right) \cdot [\veczero;\hat{\vecr}_i] $ \\[0.75em]
    \tikzcircle[black, fill={rgb,255:red,255; green,192; blue,0}]{0.5ex}
    Cable Tension & $\mathcal{F}_i = t\left[ \vecb_i\times \hat{\vecr_i}, \hat{\vecr}_i \right]$ \\
    \tikzcircle[fill={rgb,255:red,255; green,0; blue,0}]{0.5ex}
    Winch & $\tau_i = \left(t_i r - \mathcal{I} \ddot{l}_i / r + f(l_i, \dot{l}_i) r \right)$ \\
    \tikzcircle[black, fill={rgb,255:red,127; green,127; blue,127}]{0.5ex}
    Wrench & $ \sum_i [\mathrm{Ad}_{[\veczero;\vecb_i]}]^T\mathcal{F}_i + mg = \mathcal{G}\dot{\mathcal{V}} - [\mathrm{ad}_\mathcal{V}]^T \mathcal{G}\mathcal{V}$ \\[.3em]
    \tikzcircle[red, fill={rgb,255:red,0; green,0; blue,0}]{0.5ex}
    Collocation & $T[k+1] = T[k] + \mathrm{dt}\mathcal{V}[k]$ \\
      & $\mathcal{V}[k+1] = \mathcal{V}[k] + \mathrm{dt}\dot{\mathcal{V}}[k]$
  \end{tabular} \vspace*{0.5em}\\
  \end{center}
  where
\begin{itemize}
  \item $T_d$ is the desired pose;
  \item $Q$ is the state objective cost matrix from iLQR;
  \item $R$ is the control cost matrix from iLQR;
  \item $\tau_{min}$ and $\tau_{max}$ are the minimum and maximum allowable torques respectively (based on \cite{Pott09ckin_forcedist_closedform});
  \item $T_0$ and $\mathcal{V}_0$ are the initial pose and twist respectively; and
  \item $\veca_i$ is the routing pulley location (on the frame);
  \item $\vecb_i$ is the cable mounting point on the end effector (in the end effector frame);
  \item $\vecr_i = T\vecb_i-\veca_i$ is the vector from cable $i$'s routing pulley location to end effector mounting point;
  \item $\hat{\vecr}_i = \frac{\vecr_i}{\norm{\vecr_i}}$ is the normalized $\vecr_i$;
  \item $\mathrm{Ad_T}$ is the Adjoint of the transformation $T$ \cite{Lynch17book_modern_robotics};
  \item $\mathrm{ad}_\mathcal{V}$ is the adjoint of the twist $\mathcal{V}$;
  \item $f(l, \dot{l}) = - (\mu \tanh(50\dot{l}) + b \dot{l})$ is the friction \cite{Gouttefarde15TRO_CableControl};
  \item $\mathcal{G}$ is the inertia matrix;
  \item $\mathrm{dt}$ is the time step duration.
\end{itemize}
  The second-order effects for the cable acceleration kinematics were assumed to be negligible.  Time indices $[k]$ are omitted for readability, except for the initial state and collocation factors.
  \label{tab:factorequations}
\end{table}

We use GTSAM to optimize the factor graph.  For the elimination order, we eliminate timesteps using the ordering $K, K-1, \ldots,1$ and, within each timestep, eliminate variables using the ordering $l, \dot{l}, \ddot{l}, \dot{\mathcal{V}}, \mathcal{F}, t, \tau, \mathcal{V}, T$ (the order of cable index subscripts $i$ is arbitrarily chosen to be ascending).  The optimization result directly gives, for each timestep, $\vecx_\ff$ as the translational components of $[T^*; \twist^*]$ and $\vecu_\ff$ as $\tau^*$, where $\cdot^*$ denotes the optimized value.  From the Bayes Net resulting from the final iteration's elimination step, for each timestep the conditional probability distribution of $\tau$ is given in the form:
$$
  p(\tau ~|~ \twist, T) \propto \exp \left(\norm{R_\tau \Delta\tau - S_{\tau, \twist}\Delta\twist-S_{\tau,T}\Delta T}^2_\Sigma\right)
$$
where $R_\tau$ is an upper triangular matrix, $S_{\tau,\twist}$ and $S_{\tau,T}$ are matrices, $\Sigma$ is a covariance matrix (which bears a relationship to the value function but is unused), $\Delta\square$ denotes the difference $\square-\square^*$, and the time indices $[k]$ have been omitted for readability.
\pagebreak
\\
Then the feedback law is:
\begin{align}
  \Delta\tau^*(\Delta \twist, \Delta T) &= (R_\tau^{-1}S_{\tau,\twist}) \Delta\twist + (R_\tau^{-1}S_{\tau,T})\Delta T \label{eq:deltaT_conditional}
\end{align}
where again the time indices $[k]$ are omitted for readability.
Since $R_\tau$ is triangular, $R_\tau^{-1}S$ is trivial to compute with backsubstitution.

Finally, rewriting \eqref{eq:deltaT_conditional} in the form of \eqref{eq:online_controller}, we can now write the control law and gain matrix $K$ for each timestep:
\begin{align*}
  \vectau = ~&\Delta \tau^*(\Delta \twist, \Delta T) + \tau^*\\
  =~&K'\begin{bmatrix}\Delta\twist\\\Delta T\end{bmatrix} + \vecu_\ff\\
  K' = ~&\begin{bmatrix} R_\tau^{-1}S_{\tau,\twist} \\ R_\tau^{-1}S_{\tau,T}\end{bmatrix}
\end{align*}
where $K$ is taken from columns $1,2,4,5$ of $K'$ (the rotational feedback terms are dropped), and again the time indices $[k]$ are omitted for readability.

As described in Section \ref{sssec:software_approach_control}, to be used in \eqref{eq:online_controller}, $K$ and $\vecu_\ff$ are interpolated using a zero-order hold, while $\vecx_\ff$ is interpolated using a first-order extrapolation from the most recent discrete $\vecx_\ff$.

}{}

\bibliographystyle{IEEEtran}
\bibliography{nsf/refs,nsf/Dellaert,nsf/painting,nsf/Sang,nsf/Gerry,nsf/seth-refs,gerry_Feb2021,Gerry}

\end{document}